\documentclass[sigconf]{acmart}
\usepackage{algorithmic}
\usepackage[ruled,linesnumbered]{algorithm2e}
\usepackage{multirow}
\usepackage{color}
\usepackage[normalem]{ulem}
\usepackage{bm}
\setcitestyle{numbers,sort&compress}

\def \II{{\mathbb{I}}}
\def \x{{\bm{x}}} %
\def \X{{\mathcal{X}}} %
\def \D{{\mathcal{D}}} %
\def \J{{\mathcal{J}}} %
\def \Y{{\mathcal{Y}}} %
\def \L{\mathcal{L}} %
\def \S{\mathcal{S}} %
\def \B{\mathcal{B}} %
\def \H{H} %
\def \D{D} %
\def \R{\mathbb{R}} %
\def \E{\mathcal{E}} %
\def \M{\mathcal{M}} %

\def\etal{\emph{et al}.}
\AtBeginDocument{%
  \providecommand\BibTeX{{%
    \normalfont B\kern-0.5em{\scshape i\kern-0.25em b}\kern-0.8em\TeX}}}

\copyrightyear{2021}
\acmYear{2021}
\setcopyright{acmlicensed}\acmConference[MM '21]{Proceedings of the 29th
ACM International Conference on Multimedia}{October 20--24, 2021}{Virtual
Event, China}
\acmBooktitle{Proceedings of the 29th ACM International Conference on
Multimedia (MM '21), October 20--24, 2021, Virtual Event, China}
\acmPrice{15.00}
\acmISBN{978-1-4503-8651-7/21/10}
\acmDOI{10.1145/3474085.3475296}

\begin{document}
\fancyhead{}

\title{MGH: Metadata Guided Hypergraph Modeling for Unsupervised Person Re-identification}

\author{Yiming Wu}
\affiliation{
  \institution{Zhejiang University}
  \city{Hangzhou}
  \state{Zhejiang}
  \country{China}}
\email{yimingwu0@gmail.com}

\author{Xintian Wu}
\affiliation{
  \institution{Zhejiang University}
  \city{Hangzhou}
  \state{Zhejiang}
  \country{China}}
\email{hsintien@zju.edu.cn}

\author{Xi Li}
\authornote{Corresponding author}
\affiliation{
  \institution{Zhejiang University}
  \city{Hangzhou}
  \state{Zhejiang}
  \country{China}}
\email{xilizju@zju.edu.cn}

\author{Jian Tian}
\affiliation{
  \institution{Zhejiang University}
  \city{Hangzhou}
  \state{Zhejiang}
  \country{China}}
\email{tianjian29@zju.edu.cn}

\renewcommand{\shortauthors}{Wu, et al.}

\begin{abstract}
    As a challenging task, unsupervised person ReID aims to match the same identity with query images which does not require any labeled information.
    In general, most existing approaches focus on the visual cues only, leaving potentially valuable auxiliary metadata information (e.g., spatio-temporal context) unexplored.
    In the real world, such metadata is normally available alongside captured images, and thus plays an important role in separating several hard ReID matches.
    With this motivation in mind, we propose~\textbf{MGH}, a novel unsupervised person ReID approach that uses meta information to construct a hypergraph for feature learning and label refinement.
    In principle, the hypergraph is composed of camera-topology-aware hyperedges, which can model the heterogeneous data correlations across cameras.
    Taking advantage of label propagation on the hypergraph, the proposed approach is able to effectively refine the ReID results, such as correcting the wrong labels or smoothing the noisy labels. 
    Given the refined results, We further present a memory-based listwise loss to directly optimize the average precision in an approximate manner. 
    Extensive experiments on three benchmarks demonstrate the effectiveness of the proposed approach against the state-of-the-art.
\end{abstract}

\ccsdesc[500]{Information systems~Image search}
\keywords{Unsupervised Person Re-Identification; Metadata; Hypergraph; Listwise Loss; Memory}

\maketitle

\section{Introduction}

\begin{figure}[t]
    \centering
    \includegraphics[width=0.95\linewidth]{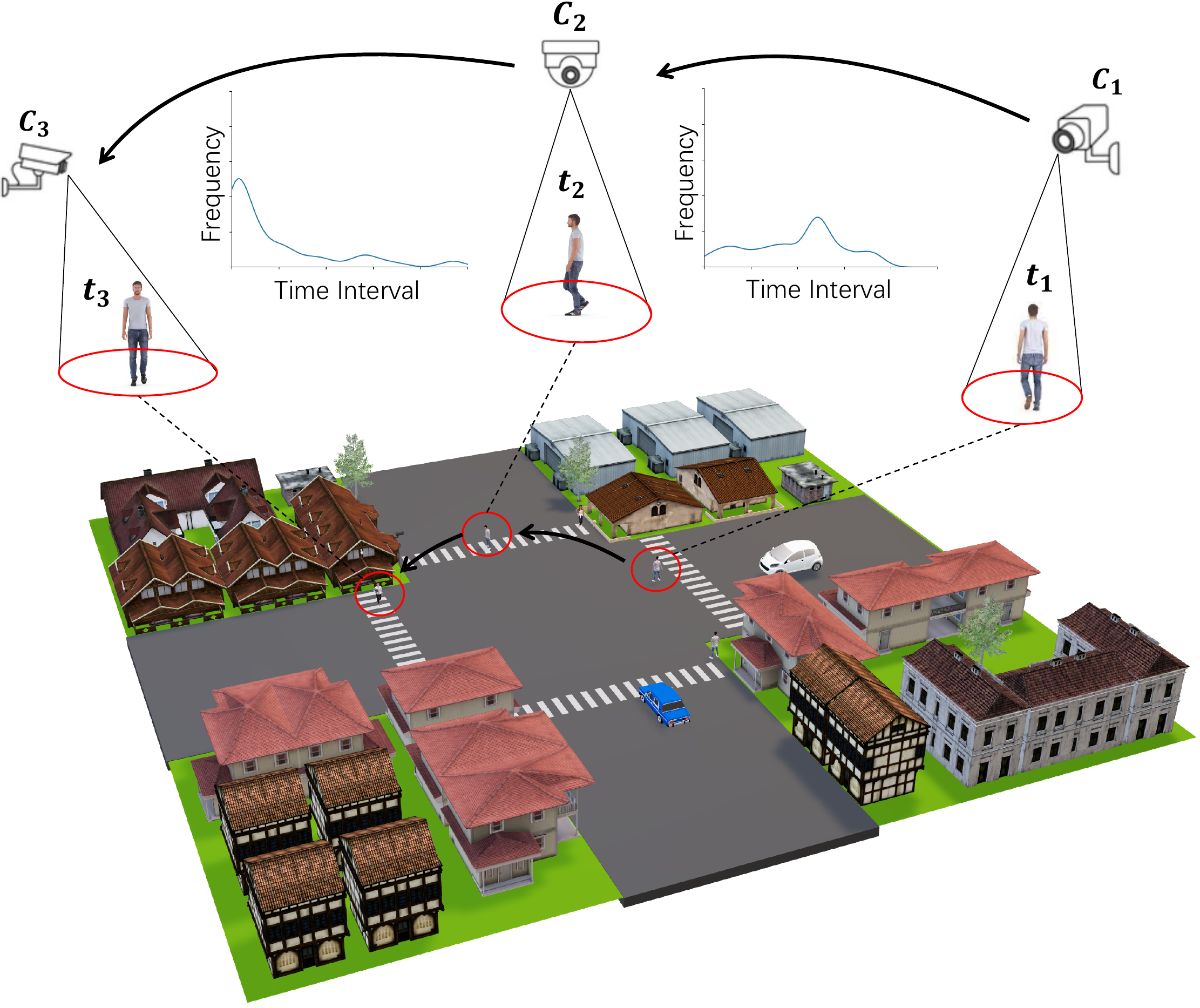}
    \caption{Illustration of the distributed camera network. The metadata (i.e. camera index, timestamp, etc.) is valuable to distinguish as well as associate persons of interest. Camera information is helpful to guide the learning of camera-invariance feature representation. Camera-timestamp tuples could be used to measure the transition probability with a prior spatio-temporal distribution.}
    \Description{Motivation of leveraging metadata.}
    \label{fig:camera network}
\end{figure}

Given a person of interest for query, person re-identification(ReID) aims to search the same identities against gallery, it has been widely used in many real-world applications, such as video surveillance system, robotics, human-computer interaction, etc. 
Due to the expensive annotation cost, 
recent researches focus on the unsupervised person ReID, 
which requires no manual annotations. Moreover, ReID is generally carried out in a distributed camera network, where a wealth of auxiliary metadata (e.g. camera index, timestamp, etc.) is attached with the captured images. As shown in Figure~\ref{fig:camera network}, the metadata could provide auxiliary guidance to unsupervised person ReID~\cite{wang2020camera,wu2019unsupervised,lv2018unsupervised,wang2019spatial,li2020joint,xie2020progressive,liao2020interpretable,zhong2020learning,zhong2018generalizing,luo2020generalizing}, but how to adequately utilize such heterogeneous structure in unsupervised ReID is still an open problem.

\begin{figure}[t]
    \centering
        \includegraphics[width=0.9\linewidth]{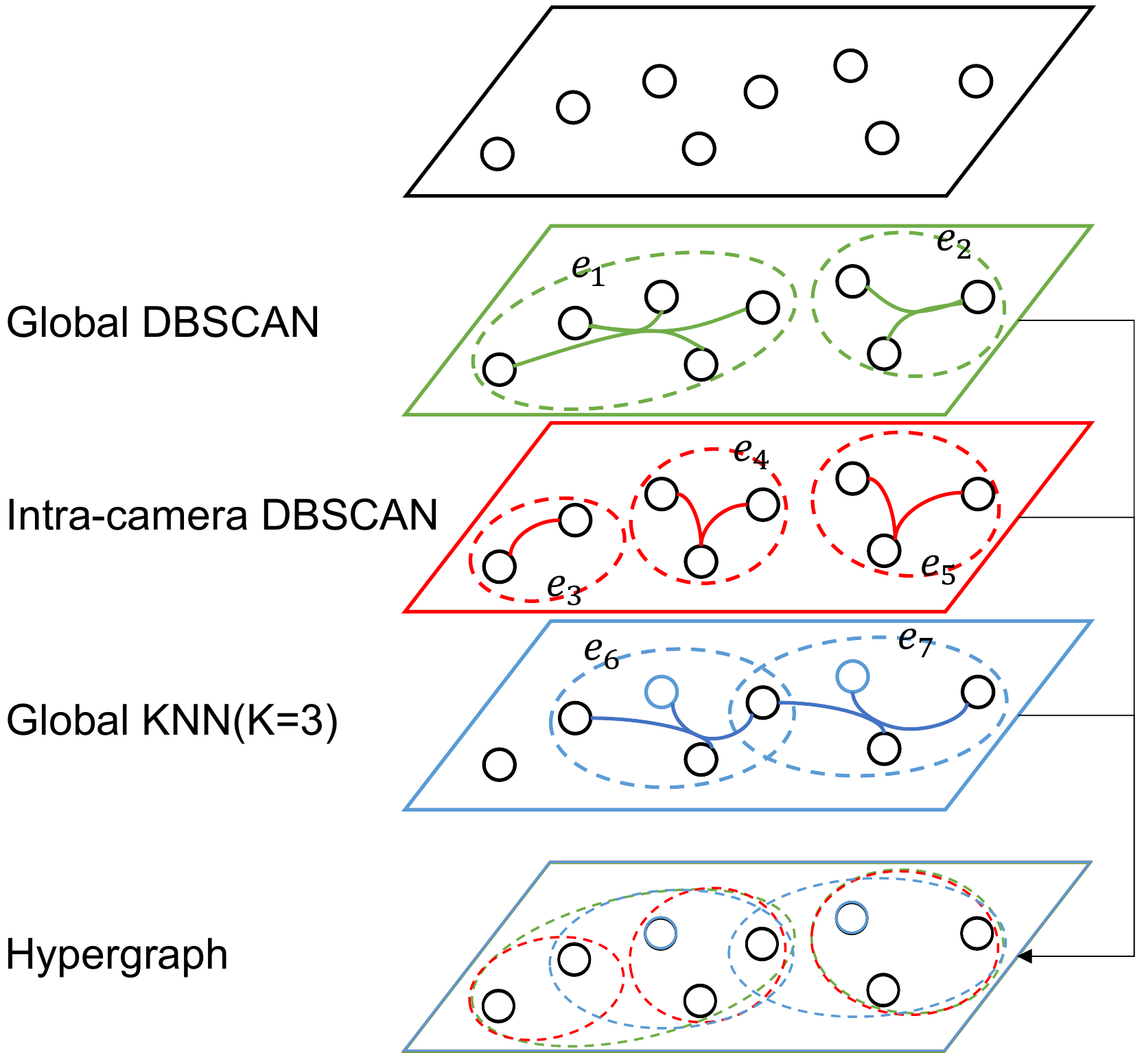}
    \caption{Illustration of hypergraph construction. We investigate different hyperedge construction strategies, i.e. KNN, clustering, camera-aware clustering. 
    Global Clustering: perform global clustering and group the instances in a cluster as a hyperedge. Intra-camera Clustering: camera information is combined with clustering to perform intra-camera Clustering. Global KNN: given a vertex (i.e., centroid), a hyperedge connects itself and its nearest neighbors.}
    \Description{How to construct hyperedges.}
    \label{fig:hypergraph}
\end{figure}

Recently, unsupervised person ReID approaches~\cite{ge2020self,fu2019self,zheng2020exploiting} follow clustering-and-finetune pipeline, which iteratively assigns pseudo labels for data then trains the feature extractor with the pseudo labels. In practice, since the ReID image data is generated from a distributed camera network,
the correlation among images is usually diverse, heterogeneous, and complicated. 
Therefore, the relations are various, such as visual connections and camera connections, which causes the unreliable pseudo labels in unsupervised person ReID. To solve this issue, a hypergraph representation~\cite{huang2010image} is an appropriate tool to model the multi-modal and heterogeneous data correlation by means of a variety of hyperedges~\cite{gao2020hypergraph}.
Due to the heterogeneous information in the metadata, the relationship among instances are varied from the different viewpoints. 
To address this issuse, we propose three kinds of hypergraph construction techniques: ``Global Clustering'', ``Intra-camera Clustering'', and ``Global KNN''. As illustrated in Figure~\ref{fig:hypergraph}, ``Global Clustering'' means performing DBSCAN on all data and then grouping the instances in a cluster as a hyperedge, this strategy aims to capture the global density-based mode in the data. ``Intra-camera Clustering'' considers the camera information and performs DBSCAN under an individual camera, this strategy concentrates on capture the camera-conditioned marginal density mode. ``Global KNN'' stands for building a hyperedge by connecting one vertex with its $K$ nearest neighbors, this strategy focuses on local neighborhood to capture the smoothness locally. By means of merging several hyperedges created by different approaches, the hypergraph is able to perceive the camera topology and model the heterogeneous data correlation across cameras. With this hypergraph, we can generate more reliable pseudo labels in learning process.

Given the generated pseudo labels, the ReID model could be finetuned iteratively. Several approaches related to mutual learning~\cite{ge2020mutual,wang2020attentive,chen2021enhancing,zhao2020unsupervised} and memory-based contrastive loss~\cite{wang2020camera,zhong2020learning} are proposed to learn the robust feature representation with noisy labels. Although these proposed classification losses are verified to be effective, the end task such as average precision is not directly optimized in these approaches. To help address this issue, we propose a listwise loss combined with an instance-level memory, which provides a fine-grained supervision by directly optimize average precision in an approximate manner. 
Besides, we combine the proposed listwise loss and camera-aware contrastive loss as coarse-to-fine supervision for model updating.

Based on the motivation above, we propose a novel method termed \textbf{M}etadata \textbf{G}uided \textbf{H}ypergraph (MGH) to mutually guide the process of label refinement and feature learning. Specifically, we construct a heterogeneous hypergraph based on the metadata to generate pseudo labels, and utilize coarse-to-fine memory-based supervision to update the model. To achieve this, we first generate the noisy pseudo labels by clustering the visual feature representations of all unlabeled images. Then utilize hypergraph for label refinement. Specifically, we construct the hyperedges based on a joint similarity matrix~\cite{wang2019spatial} by considering visual information and spatio-temporal context simultaneously. Hyperedges are grouped to generate a hypergraph, which models the complicated high-order relationships among the data. Then, we perform label propagation on the hypergraph structure, which rectifies the label errors caused by the previous clustering algorithm. For model updating, the instance-level memory and camera-aware prototype memory are constructed to simultaneously capture local and global distribution. We propose to use a listwise loss based on the instance-level memory, which is combined with a camera-aware contrastive loss to provide coarse-to-fine supervision.

In summary, our main contributions are three-fold as follows:
\begin{itemize}
    \item We propose a heterogeneous hypergraph to model the complicated data correlation among the metadata, which facilitates the unsupervised person ReID in the real-world scenario.
    \item We propose a novel unsupervised person ReID model named \textbf{MGH}, which consists of label generation with hypergraph and model updating through memory-based coarse-to-fine supervision.
    \item On three public person ReID benchmarks with readily available metadata, i.e. camera index and timestamp, our proposed method outperforms the state-of-the-art approaches.
\end{itemize}

\section{Related Work}

\subsection{Person ReID}
\noindent \textbf{Unsupervised Person ReID.} Recent deep learning-based unsupervised person ReID and highly correlated unsupervised domain adaptive person ReID approaches could be classified into two campuses: (1) distribution alignment, and (2) pseudo-label-based methods. For distribution alignment approaches, the purpose is to learn domain invariant feature representations. In~\cite{deng2018image,zhong2018generalizing,zheng2019joint,liu2020unity,huang2020real,zheng2017unlabeled,luo2020generalizing,zou2020joint,jin2020style}, generative models such as a generative adversarial network (GAN) are exploited to achieve image-to-image translation from the source domain to the target domain and then use the generated images to train the model. Some other approaches~\cite{mekhazni2020unsupervised} align the feature space by MMD loss. Current leading approaches perform label estimation and train the model accordingly. Such an operation is iteratively executed until the model converges. Fan~\etal~\cite{fan2018unsupervised} and Song~\etal~\cite{song2020unsupervised} firstly utilize such pipeline for unsupervised person ReID. SSG~\cite{fu2019self} extends these methods by utilizing clustering on the part features and global features. BUC~\cite{lin2019bottom} and HTC~\cite{zeng2020hierarchical} propose bottom-up clustering algorithm tailored for person ReID. 
Besides, in~\cite{wang2020unsupervised}, memory-based multi-label classification loss named MMCL is proposed, which combines non-parametric classifier, multi-label classification, and single-label classification in a unified framework. Similarly, Lin~\etal~\cite{lin2020unsupervised} use a soft label to represent instance similarity with custom distance metric for unsupervised person ReID. Furthermore, in~\cite{chen2021enhancing,zheng2020exploiting,feng2021complementary,zhai2020multiple,ge2020mutual,zhai2020ad,zhao2020unsupervised}, mutual learning combines the off-line hard pseudo label and on-line soft pseudo labels in an alternative training scheme.

\vspace{0.5em}
\noindent \textbf{Person ReID with Metadata.}
As mentioned above, there is a wealth of valuable meta information in the distributed camera network, such as camera index, timestamp, etc. Wang~\etal~\cite{wang2019spatial} propose a joint metric considering spatio-temporal information and appearance similarity for supervised person ReID. Lv~\etal~\cite{lv2018unsupervised} propose an unsupervised incremental learning scheme named TFusion, which builds up a Bayesian fusion model to combine the spatio-temporal pattern with visual features. Liao~\etal~\cite{liao2020interpretable} propose an effective post-processing strategy named TLift to enhance the retrieval stability of ReID model. In~\cite{wu2019iccvunsupervised,wang2020camera}, camera-aware learning is proposed to simultaneously consider intra-camera consistency and inter-camera matching.

\subsection{Hypergraph}
Graphs have been widely explored in computer vision tasks, such as image classification~\cite{chen2019multi}, image retrieval~\cite{huang2010image}, person ReID~\cite{wu2020adaptive}, segmentation~\cite{ji2020context} etc. While the conventional graph is validated effective in the recent approaches, it could only model the pairwise relationships, which is hard to extend to the complicated data structure. Therefore, hypergraph~\cite{zhou2006learning} is introduced to model the high-order data correlation. An~\etal~\cite{an2016person} utilize hypergraph to learn the weight for multi-modal feature fusion in person ReID. Besides, Yan~\etal~\cite{yan2020learning} leverage hypergraph neural network to learn the multi-granular feature representation for video-based person ReID. Different from these methods, we utilize hypergraph for label refinery. By propagating residual label error on the hypergraph and smoothing the corrected labels, we obtain the more reliable labels.

\subsection{Metric Learning}
In the computer vision community, metric learning~\cite{liu2011learning,ji2019human,wang2018progressive,zhao2021video} is widely used for learning discriminative feature representations. In face recognition, classification loss such as margin-based cross entropy loss~\cite{wang2018additive} is widely explored. In~\cite{hermans2017defense}, triplet loss is widely studied for person ReID, and recent approaches~\cite{luo2019bag,wang2018learning} for person ReID commonly combine cross entropy loss and triplet loss to train the ReID model. However, these objective functions could not directly optimize the ranking tasks. In image retrieval, the listwise loss is widely explored to optimize the mean average precision (mAP). Inspired by the recent works~\cite{wang2020cross,brown2020smooth}, we develop a memory-based listwise loss to directly optimize AP in an approximate manner.

\section{Methodology}\label{sec:method}
\begin{figure*}[t]
  \begin{center}
  \includegraphics[width=0.95\linewidth]{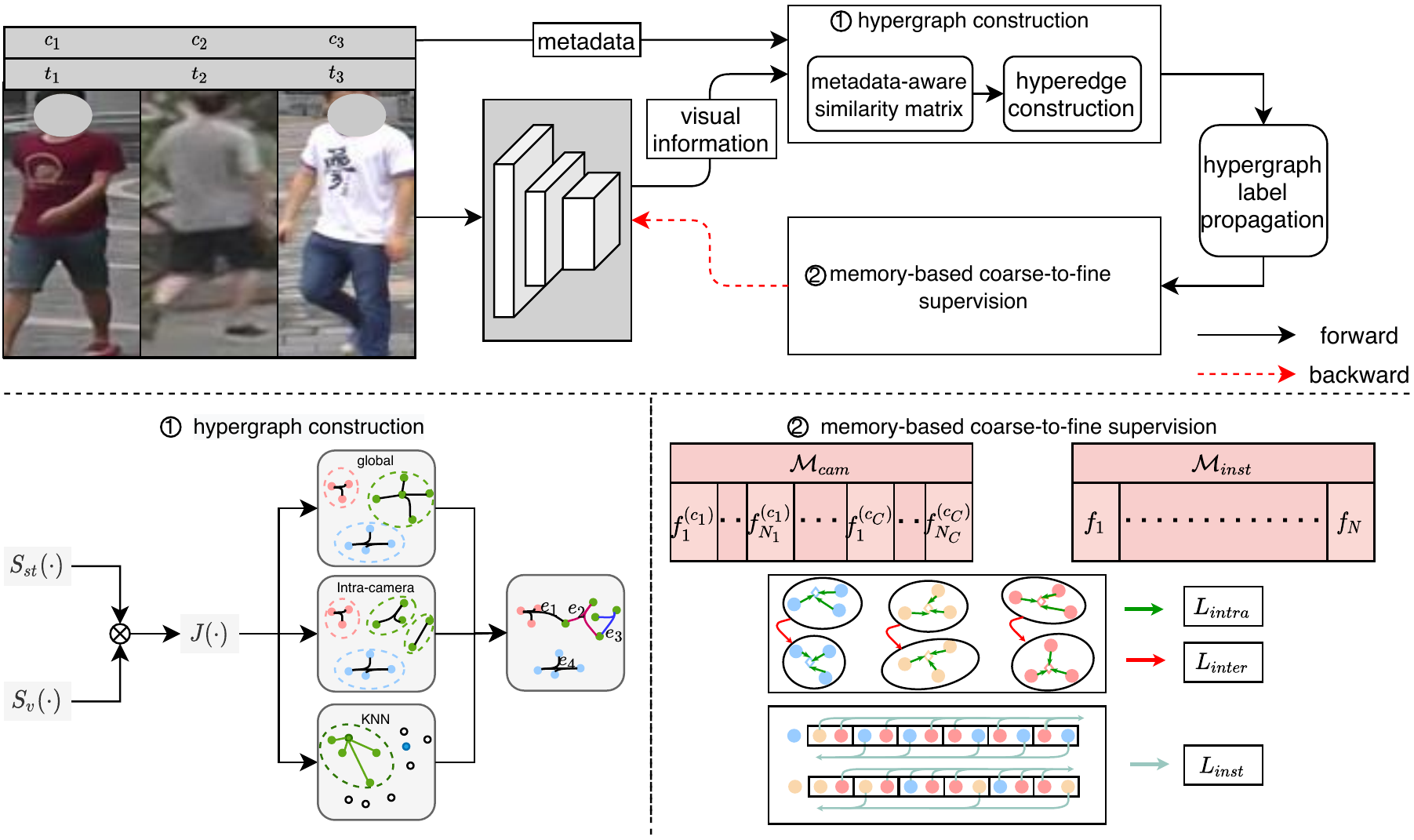}
  \end{center}
  \caption{Diagram of our proposed method MGH. Our proposed method is a pseudo-label-based method, which follows the clustering-and-finetune pipeline that iteratively assigns the pseudo label for data and then finetunes the feature extractor until the training converges.}
  \Description{The pipeline of our proposed method.}
  \label{fig:MGH}
\end{figure*}

Unsupervised person ReID aims to adapt the pretrained model to the target domain, which contains plenty of images without annotations but rich metadata such as camera index and timestamp. In this paper, we apply metadata including camera index and timestamp to improve the performance of ReID model. %

Given an unlabeled target dataset $\mathcal{D}=\{(\x_i, c_i, t_i)\}_{i=1}^N$ with $N$ images,
where $\x_i$ is $i$-th image, $c_i, t_i$ are its camera index and timestamp respectively.
We adopt a hypergraph $G=(V, E, w)$ to encode the complex relationships, 
where $V$, $E$, and $w$ are vertex set, hyperedge set, and hyperedge weights, respectively.
The hyperedge is built based on the similarity matrix $\J$,
where $\J(i, j)$ measures the pairwise similarity for vertex $i$ and $j$.
With the hypergraph, we perform label propagation to generate robust pseudo labels.
Furthermore, we could fine-tune the ReID model $f_\theta$ with parameters $\theta$, which is initialized by pretrained model.
Table~\ref{tab:notation table} shows the main notations used here in after.

\begin{table}[htbp]
    \centering
    \begin{tabular}{c c}
        \toprule
        Symbol            & Brief Description \\
        \hline
        $\x, c, t$        & image, camera index, timestamp \\
        $\mathcal{D}$     & dataset  \\
        $N$               & number of images \\
        $f_{\theta}$      & ReID model \\
        $\J$             & joint similarity matrix \\
        $G, V, E$        & hypergraph, vertices, hyperedge \\
        $w, \H $         & hypergraph weight, incidence matrix  \\
        $d(v), \delta(e)$ & vertex, hyperedge degree \\
        $\D_v, \D_e$     & vertex, hyperedge degree matrix\\
        $W$              & hyperedge weight matrix\\
        $\S_v, \S_{st}$  & visual and spatio-temporal similarity matrix  \\
        $A$              & hypergraph adjacency matrix \\
        $\Delta_t$       & time interval in histogram\\
        $N_c$            & number of cluster \\
        $n_{c_i,c_j}^b$  & number of image pairs with the same labels \\
        $\B$             & histogram  \\
        $y, \Y$          & pseudo label \\
        $\hat{\Y}$       & corrected pseudo label \\
        $\E_r, \E_u$ & error matrix \\
        $\mathbb{S}_r, \mathbb{S}_u$ & instance set \\
        $\Y_{soft} $     & soft pseudo label \\
        \bottomrule
    \end{tabular}
    \caption{Notation table.}
    \label{tab:notation table}
\end{table}

\vspace{-1.5em}
\subsection{Meta Hypergraph Construction}
\vspace{0.5em}
\noindent\textbf{Meta Hyperedge Construction}\\
Metadata information is crucial for modeling the high-order heterogeneous data correlation in a distributed camera network, and such data correlation plays a vital role in distinguishing hard ReID matchings in the real-world scenario. To well model the diverse, heterogeneous, and complicated correlation, hypergraph is an appropriate tool by utilizing flexible hyperedges. How to define the hypergraph directly affects the effectiveness of the data correlation modeling. As illustrated in Figure~\ref{fig:MGH}, we propose three kinds of hyperedge construction approaches, termed ``Global Clustering'', ``Intra-camera Clustering'', and ``Global KNN'' respectively, the different hyperedges could complement each other to encode complicate high-order data correlations.

\begin{itemize}
    \item \textbf{Global Clustering}: we use clustering to directly group all vertices into clusters, and connects the vertices in each cluster by a hyperedge.
    \item \textbf{Intra-camera Clustering}: with the camera information, all vertices could be separated by the different camera, then we perform clustering to group vertex in each camera.
    \item \textbf{Global KNN}: in this method, we build a hyperedge by connecting one vertex with its $K$ nearest neighbors.
\end{itemize}
By merging several hyperedges created by different approaches, the hypergraph is able to perceive the camera topology and model the heterogeneous data correlation across cameras.
Besides, the weight of each hyperedge can be measured by:
\begin{equation}
    w(e)=\sum_{i, j \in e} exp\left( \frac{\J(i, j)^2}{\sigma^2} \right),
    \label{eq:hyperedge weight}
\end{equation}
where $(i, j)$ denote a pair of vertices in the hyperedge $e$, and 
$\sigma$ is the medium value of the similarity of all vertex pairs.
Then the hypergraph could be represented by an incidence matrix $\H\in \R^{|V|\times|E|}$:%
\begin{equation}
    \H(v_i, e_j) = 
    \begin{cases}
      \J(j, i) & \text{if $v_i \in e_j$}\\
      0 & \text{otherwise}.
    \end{cases}
    \label{eq:incidence matrix}
\end{equation}

According to this definition, higher weighted hyperedge includes the vertices with higher inner similarity. 
Besides, the vertex degree and hyperedge degree are defined as $d(v)=\sum_{e\in E}w(e)\H(v,e)$ and $\delta(e)=\sum_{v\in V}\H(v,e)$, respectively. 
The corresponding diagonal matrices of the vertex degree, the hyperedge degree, and hyperedge weight are $\D_v$, $\D_e$, and $W$, respectively. 
Then we define the hypergraph adjacency matrix as follows:
\begin{equation}
    A=\H W \D_e^{-1} \H^{\top}.
\end{equation}
With the adjacency matrix, we could perform label propagation to correct the label noise.

\vspace{0.5em}
\noindent\textbf{Metadata-aware Similarity Matrix}~\label{sec:metadata-aware similarity metric}\\
To construct the hyperedges, we present the calculation of joint similarity matrix $\J$.
As shown in Figure~\ref{fig:MGH}, 
by jointly considering the visual similarity and spatio-temporal similarity, the joint pairwise similarity is 
\begin{equation}
    \J(i, j) = \frac{1}{1+\lambda_0e^{-\gamma_0S_v(i,j)}}\cdot\frac{1}{1+\lambda_1e^{-\gamma_1S_{st}(i,j)}},
    \label{eq:joint similarity}
\end{equation}
where $S_v(i, j)$ and $S_{st}(i,j)$ are the visual and spatio-temporal similarity of the images $\x_i$ and $\x_j$, respectively, 
and $S_v(i,j)$ is cosine similarity in our implementation.

To obtain the spatio-temporal similarity,
we first perform clustering based on features $\{f_\theta(\x_i)\}_{i=1}^N$ and get the noisy pseudo labels $\Y=\{y_i\}_{i=1}^N$, where $y_i \in \R^{N_c}$ and $N_c$ is the number of clusters. 
With the pseudo labels, we could build the spatio-temporal pattern through the spatio-temporal distribution,
which is represented as the normalized histograms:
\begin{equation}
    \B(\II(y_i,y_j) | c_i, c_j, b)=\frac{n^b_{c_i,c_j}}{\sum_l n^l_{c_i,c_j}}, 
    \label{eq:histogram}
\end{equation}
where $b$, $c$, and $\II(\cdot)$ are the index of histogram bins, the camera index, and indicator function, respectively. And $n^b_{c_i,c_j}$ is the number of image pairs with the same labels whose time differences are in the scale of $((b-1)\Delta_t, b\Delta_t)$.
Based on this spatio-temporal distribution, we can estimate the spatio-temporal consistency of two images $(\x_i, c_i, t_i)$ and $(\x_j, c_j, t_j)$ as follows:
\begin{equation}
    \S_{st}(i, j) = \B(\II(y_i,y_j) | c_i, c_j, \lfloor \frac{|t_i - t_j|}{\Delta_t}\rfloor ),
    \label{eq:spatio-temporal similarity}
\end{equation}
where $\lfloor\cdot\rfloor$ is round function.

\subsection{Meta Hypergraph Guided Label Generation}~\label{sec:label generation}\\
In unsupervised person ReID, the pseudo-label-based methods generally follow the pipeline that alternatively assigns pseudos label and finetune the feature extractor until the training converges. Given the extracted feature representations, the first step is to generate reliable pseudo labels, which means grouping the same identities into an individual cluster. In the previous methods, several clustering algorithms such as K-Means and DBSCAN~\cite{song2020unsupervised,Lin2020,NEURIPS2020_f3ada80d} have been explored.
The existing methods generally perform label generation based on the pairwise visual relationships, which neglect the auxiliary spatio-temporal constraints. 
Here we adopt a hypergraph to model the diverse and heterogeneous data structure, and then refine the noisy pseudo label by either correcting the wrong labels or smoothing the noisy labels on a hypergraph.

Label propagation~\cite{zhou2004learning,zhang2019multiple,jia2020residual,klicpera2018combining} has been adopted to smooth the label prediction. Following the motivation that label prediction is highly correlated along edges in the graph, we perform label propagation on the hypergraph for label refinement. 

Specifically, we first select 4 instances each cluster to be the most reliable instances $\mathbb{S}_r$, and the reliable prediction $\Y_r$, the extra unreliable instances are defined as $\mathbb{S}_u$. Then we could define error matrix $\E$ as follows:
\begin{equation}
    \E_r = \Y - \Y_r, \E_u = 0,
    \label{eq:error matrix}
\end{equation}
where $\Y_r \in \R^{N\times N_c}$ consists of one-hot vector on the selected reliable predictions and zero elsewhere, $\Y$ is the initial noise label, and $\E_u$ is the error matrix for unreliable instances. Then we perform iterative label propagation upon the hypergraph following the scaled fixed diffusion method proposed by~\cite{huang2020combining}, which iteratively propagates residual error as follows:
\begin{equation}
    \E_u^{(t+1)}=\alpha_1[D^{-1/2}AD^{-1/2}\E^{(t)}]_u,
    \label{eq:residual label correction}
\end{equation}
where $[\cdot]_u$ means selecting unreliable entries in the error matrix.
We fix reliable prediction $\E_r^{(t)}=\E_r$ until converging to $\hat{\E}$, and obtain the corrected label prediction $\hat{\Y}_r=\Y_r+s\hat{\E}$, where $s$ is a scale coefficient. Further, we smooth the prediction by iterative label propagation as follows:
\begin{equation}
    \Y^{(t+1)}=(1-\alpha_2)\Y+\alpha_2 D^{-1}A\Y^{(t)}.\
    \label{eq:label smooth}
\end{equation}
The iteratively updated prediction $\Y^{t}$ converges to final prediction.%

\begin{algorithm}[t]
    \SetAlgoLined
    \KwIn{Unlabeled dataset $\D=\{(\x_i, c_i, t_i)\}_{i=1}^N$, pretrained ReID model $f_\theta$, number of epochs $Epoch$, number of iterations in each epoch $Iters$.}
    \KwOut{Finetuned ReID model $\hat{f_{\theta}}$.}
    \While{$1 \leq$ i $\leq $ $Epoch$}{
        \tcc{Meta Hypergraph Guided Label Generation}
        Construct joint similarity matrix based on Equ.~\ref{eq:joint similarity};\\
        Construct incidence matrix $\H$ based on Equ.~\ref{eq:incidence matrix};\\
        Generate pseudo label $\hat{\Y}$ based on Equ.~\ref{eq:residual label correction} and Equ.~\ref{eq:label smooth};\\
        \tcc{Memory-based Coarse-to-fine Supervision}
        Construct instance-level memory $M^{inst}$ and camera-aware prototype memory $M^{cam}$;\\
        \For{j = $1$:$Iters$}{
            Calculate entire loss by Equ.~\ref{eq:loss};\\
            Update ReID model $f_{\theta}$;
        }
     }
    Output finetuned ReID model $\hat{f_{\theta}}$.
    \caption{Meta Hypergraph Guided Network}
    \label{alg:algorithm}
\end{algorithm}

\subsection{Memory-based Coarse-to-fine Supervision}\label{sec:memory-based supervision}
After label correction with hypergraph label propagation, we obtain the reliable labels $\hat{\Y}$. We propose to utilize memory-based coarse-to-fine supervision to guide the model training. More specifically, we build two memory banks: instance-level memory $\M^{inst}$ and camera-aware prototype memory $\M^{cam}$, which are initialized with extracted features every epoch and updated in a moving average manner during the training process.

In the instance-level memory, all feature representations are stored and updated in the training process. Based on the instance-level memory bank, we adopt a listwise loss named Smooth-AP~\cite{brown2020smooth} to directly optimize mean average precision. Different from Smooth-AP that treats the in-batch images as the gallery, we compute the relevance score between query and features in the memory bank. This makes more hard negative samples participate in loss calculation. For a query image $\x_q$, the smoothed average precision is defined as:
\begin{equation}
    \begin{aligned}
        &AP_{q} \approx \frac{1}{\left|\mathbb{S}_{P}\right|} \sum_{i \in \mathbb{S}_{P}} \frac{1+\sum_{j \in \mathbb{S}_{P}} \mathcal{G}\left(q,j\right)}{1+\sum_{j \in \mathbb{S}_{P}} \mathcal{G}\left(q,j\right)+\sum_{j \in \mathbb{S}_{N}} \mathcal{G}\left(q,j\right)},\\
        &\mathcal{G}\left(q,j\right) = \frac{1}{1+e^{-f_{\theta}{(\x_q)}^{\top} \M^{inst}_j}},
    \end{aligned}
\end{equation}
where $\mathbb{S}_P$ and $\mathbb{S}_N$ are positive set and negative set for query $q$, which are consists of 1000 instances with the highest similarity to query. Then $L_{inst}$ is defined as:
\begin{equation}
    \begin{aligned}
        L_{inst} &= \frac{1}{N}\sum_{q=1}^N(1 - AP_q). \\
    \end{aligned}
\end{equation}

Because of the variation between different scenes, intra-camera matching is more robust than inter-camera matching, we build camera-aware prototype memory to store the camera-aware prototypes, which depicts more precise camera-aware data distribution. For instance, the identity under different cameras has different prototypes. Moreover, we adopt weighted intra-camera loss and inter-camera loss to keep the intra-camera discrimination and encourage inter-camera matching, which is formulated as follows:
\begin{equation}
    \begin{aligned}
        \L_{intra} = -\sum_{c=1}^{C} \frac{1}{N_{c}} \sum_{\x_{i} \in \mathbb{S}_{c}} \sum_{k=1}^{Z_c} \hat{\Y}_{soft}(i,k) \log \frac{\phi(i, k)}{\sum_{k=1}^{Z_c} \phi(i,k)},\\    
        \L_{inter} = -\sum_{i=1}^{N^{\prime}} \frac{1}{|\mathcal{P}|} \sum_{p \in \mathcal{P}} \log \frac{\phi(i, p)}{\sum_{u \in \mathcal{P}} \phi(i, u)+\sum_{q \in \mathcal{Q}} \phi(i,q)},
    \end{aligned}
\end{equation}
where $\phi(i, k) = \exp (f_{\theta}(\x_{i})^{\top} \M_j^{cam} / \tau)$, $\mathbb{S}_{c}$ is the dataset of the $c$-th camera, $Z_c$ is the number of identity in $\mathbb{S}_c$, and $\mathcal{P}$, $\mathcal{Q}$ denote the index sets of positive and negative prototypes for $\x_i$, and $\Y_{soft}$ is soft label calculated based on $\J$ and $\hat{\Y}$ as follows:
\begin{equation}
    \hat{\Y}_{soft}(i, n_c) = \sum_{m\in M}\frac{\J(i, m)}{|M|}, M=\{m|\Y(m, n_c)= 1\},
    \label{eq:soft label}
\end{equation}
where $n_c \in [1, 2, ..., N_c]$ is the class index. For training the network, we formulate the entire loss function as follows:
\begin{equation}
    \L=\lambda_1\L_{intra}(\X, \hat{\Y}) + \lambda_2\L_{inter}(\X, \hat{\Y}) + \lambda_3\L_{inst}(\X, \hat{\Y}),
    \label{eq:loss}
\end{equation}
where $\L_{intra}$, $\L_{inter}$, and $\L_{inst}$ are weighted intra-camera loss, inter-camera loss, and instance loss, respectively. We summarize the overall algorithm is summarized in Algorithm~\ref{alg:algorithm}.

\section{Experiments}\label{sec:experiments}
\subsection{Experiment setting}\label{sec:experiment setting}
\noindent\textbf{Dataset and Protocal.} In this paper, we evaluate the performance on three public benchmarks Market1501~\cite{zheng2015scalable}, DukeMTMC~\cite{zheng2017unlabeled,ristani2016performance}, and MSMT17~\cite{wei2018person}. 
We follow the standard evaluation setting in the previous methods to evaluate our proposed method, cumulative matching characteristic (CMC) at Rank-1, Rank-5, Rank-10 and mean average precision (mAP) are reported for comparison.

\begin{table}
  \centering
  \begin{tabular}{lrrrr} 
  \hline
  \multirow{2}{*}{Model}                                                                       & \multicolumn{2}{l}{Market}                       & \multicolumn{2}{l}{DukeMTMC}                      \\ 
  \cline{2-5}
                                                                                               & \multicolumn{1}{l}{R1} & \multicolumn{1}{l}{mAP} & \multicolumn{1}{l}{R1} & \multicolumn{1}{l}{mAP}  \\ 
                                                                                               \hline
  \multicolumn{5}{l}{\textcolor[rgb]{0.502,0.502,0.502}{evaluation of individual components}}                                                                                                         \\ 
  \hline
  CAP                                                                                          & 91.4                   & 79.2                    & 81.1                   & 67.3                     \\
  CAP$^\star$                                                                                  & 91.5                   & 79.7                    & 81.5                   & 67.3                     \\
  \textbf{Sbase}                                                                               & 92.3                   & 80.8                    & 82.2                   & 68.7                     \\
  \textbf{Sbase} w/ $\L_{inst}$                                                                & 92.7                   & 81.3                    & 83.1                   & 69.3                     \\
  \textbf{Sbase} w/ HG                                                                         & 92.7                   & \textbf{81.8}           & 83.3                   & 70                       \\
  \begin{tabular}[c]{@{}l@{}}\textbf{Sbase w/ $\L_{inst}$}+HG\\(MGH)\end{tabular} & \textbf{93.2}                   & 81.7           & \textbf{83.6}                   & \textbf{70.2}                     \\ 
  \hline
  \multicolumn{5}{l}{\textcolor[rgb]{0.502,0.502,0.502}{impact of hyperparameter $\lambda_3$}}                                                                                                        \\ 
  \hline
  $\lambda_3$=0                                                                                & 92.3                   & 80.8           & 82.2                   & 68.7           \\
  $\lambda_3$=1                                                                                & 92.1                   & \textbf{81.3}           & 82.5                   & \textbf{69.3}            \\
  $\lambda_3$=10                                                                               & \textbf{92.7}          & \textbf{81.3}           & \textbf{83.1}          & \textbf{69.3}            \\
  $\lambda_3$=100                                                                              & 92.2                   & 81.2                    & 83                     & 69.1                     \\
  \hline
  \end{tabular}
  \caption{(1) Evaluation of individual component. Based on CAP~\cite{wang2020camera}, we propose a strong baseline denoted \textbf{Sbase} with weighted intra-camera loss. $\L_{inst}$ short for instance-level memory based listwise loss. HG means hypergraph label correction. Our final model combines $\L_{inst}$ and HG named MGH. (2) The Impact of Hyperparameter $\lambda_3$. We compare different value of $\lambda_3$.}
  \label{tab:ablation}
\end{table}

\begin{table*}
  \centering
  \begin{tabular}{lccccccccccc} 
  \hline
  \multirow{2}{*}{Method}                                              & \multirow{2}{*}{Reference} & \multicolumn{5}{c}{Market1501}                                              & \multicolumn{5}{c}{DukeMTMC}                                                  \\ 
  \cline{3-12}
                                                                      &                            & source   & R1             & R5             & R10            & mAP            & source   & R1             & R5             & R10            & mAP             \\ 
  \hline
  \multicolumn{12}{l}{\textcolor[rgb]{0.502,0.502,0.502}{Unsupervised person ReID }}                                                                                                                                                                               \\ 
  \hline
  BUC~\cite{lin2019bottom}            & AAAI19                     & None     & 66.2           & 79.6           & 84.5           & 38.3           & None     & 47.4           & 62.6           & 68.4           & 27.5            \\
  SpCL~\cite{ge2020self}              & NeurIPS20                  & None     & 87.7           & 95.2           & 96.9           & 72.6           & None     & 81.2           & \textcolor{blue}{90.3}           & 92.2           & 65.3            \\
  HCT~\cite{zeng2020hierarchical}     & CVPR20                     & None     & 80.0           & 91.6           & 95.2           & 56.4           & None     & 69.6           & 83.4           & 87.4           & 50.7            \\
  MMCL~\cite{wang2020unsupervised}    & CVPR20                     & None     & 80.3           & 89.4           & 92.3           & 45.5           & None     & 65.2           & 75.9           & 80.0           & 40.2            \\
  \hline
  \multicolumn{12}{l}{\textcolor[rgb]{0.502,0.502,0.502}{UDA person ReID }}                                                                                                                                                                                        \\ 
  \hline
  PUL~\cite{fan2018unsupervised}      & TOMM18                     & Duke     & 45.5           & 60.7           & 66.7           & 20.5           & Market   & 30.0           & 43.4           & 48.5           & 16.4            \\
  SPGAN+LMP~\cite{deng2018image}      & CVPR18                     & Duke     & 58.1           & 76.0           & 82.7           & 26.9           & Market   & 46.9           & 62.6           & 68.5           & 26.4            \\
  SSG~\cite{fu2019self}               & ICCV19                     & Duke     & 80.0           & 90.0           & 92.4           & 58.3           & Market   & 73.0           & 80.6           & 83.2           & 53.4            \\
  UDAP~\cite{song2020unsupervised}    & PR20                       & Duke     & 75.8           & 89.5           & 93.2           & 53.7           & Market   & 68.4           & 80.1           & 83.5           & 49.0            \\
  MMCL~\cite{wang2020unsupervised}    & CVPR20                     & Duke     & 84.4           & 92.8           & 95.0           & 60.4           & Market   & 72.4           & 82.9           & 85.0           & 51.4            \\
  MMT~\cite{ge2020mutual}             & ICLR20                     & Duke     & 89.5           & 96.0           & 97.6           & 73.8           & Market   & 76.3           & 87.7           & 91.2           & 62.3            \\
  SpCL~\cite{ge2020self}              & NeurIPS20                  & Duke     & 90.3           & \textcolor{blue}{96.2}           & \textcolor{blue}{97.7}           & 76.7           & Market   & \textcolor{green}{82.9}           & 90.1           & \textcolor{blue}{92.5}           & \textcolor{blue}{68.8}            \\
  MEB-Net~\cite{zhai2020multiple}     & ECCV20                     & Duke     & 89.9           & 96.0           & 97.5           & 76.0           & Market   & 79.6           & 88.3           & 92.2           & 66.1            \\
  AD-Cluster~\cite{zhai2020ad}        & CVPR20                     & Duke     & 86.7           & 94.4           & 96.5           & 68.3           & Market   & 72.6           & 82.5           & 85.5           & 54.1            \\
  DIM+GLO~\cite{liu2020domain}                     & MM20                       & Duke     &88.3           & 94.7           & 96.3           & 65.1          & Market     & 76.2           & 85.7           & 88.5           & 58.3            \\
  UNRN~\cite{zheng2020exploiting}     & AAAI21                     & Duke     & \textcolor{green}{91.9}           & 96.1           & \textcolor{green}{97.8}           & \textcolor{blue}{78.1}           & Market   & \textcolor{blue}{82.0}           & \textcolor{green}{90.7}           & \textcolor{green}{93.5}           & \textcolor{green}{69.1}            \\ 
  \hline
  \multicolumn{12}{l}{\textcolor[rgb]{0.502,0.502,0.502}{Unsupervised person ReID with meta information}}                                                                                                                                                    \\ 
  \hline
  JVTC+~\cite{li2020joint}            & ECCV20                     & None     & 79.5           & 89.2           & 91.9           & 47.5           & None     & 74.6           & 82.9           & 85.3           & 50.7            \\
  ECN++~\cite{zhong2020learning}      & TPAMI20                    & None     & 84.1           & 92.8           & 95.4           & 63.8           & None     & 74.0           & 83.7           & 87.4           & 54.4            \\
  CAP~\cite{wang2020camera}           & AAAI21                     & None     & \textcolor{blue}{91.4}           & \textcolor{green}{96.3}   & \textcolor{blue}{97.7}           & \textcolor{green}{79.2}    & None     & 81.1           & 89.3           & 91.8           & 67.3            \\
  TFusion~\cite{lv2018unsupervised}   & CVPR18                     & CUHK01   & 60.8           & 74.4           & 79.3           & -              & -        & -              & -              & -              & -               \\
  UGA~\cite{wu2019unsupervised}       & ICCV19                     & Tracklet & 87.2           & -              & -              & 70.3           & Tracklet & 75.0           & -              & -              & 53.3            \\
  JVTC+~\cite{li2020joint}            & ECCV20                     & Duke     & 86.8           & 95.2           & 97.1           & 67.2           & Market   & 80.4           & 89.9           & 92.2           & 66.5            \\ 
  \hline
  MGH(Ours)                           & -                                     & None     & \textcolor{red}{93.2} & \textcolor{red}{96.8} & \textcolor{red}{98.1} & \textcolor{red}{81.7 } & None     & \textcolor{red}{83.7 } & \textcolor{red}{92.1 } & \textcolor{red}{93.7 } & \textcolor{red}{70.2 }  \\
  MGH+(Ours)                          & -                                      & None     & 95.5 & 98.2 & 98.7 & 84.0 & None     & 91.0 & 94.2 & 95.3 & 77.2  \\
  \hline
  \end{tabular}
  \caption{Comparison with state-of-the-are methods on Market1501 and DukeMTMC. We compare our method to the state-of-the-art methods for unsupervised person ReID, UDA person ReID, and unsupervised person ReID with meta information. The top-3 best results are highlighted with \textcolor{red}{red}, \textcolor{green}{green}, and \textcolor{blue}{blue} respectively. MGH+ uses joint similarity to compute the query-gallery similarity.}
  \label{tab:comparison with Market and DukeMTMC}
\end{table*}

\begin{figure}[t]
  \centering
  \includegraphics[width=1.0\linewidth]{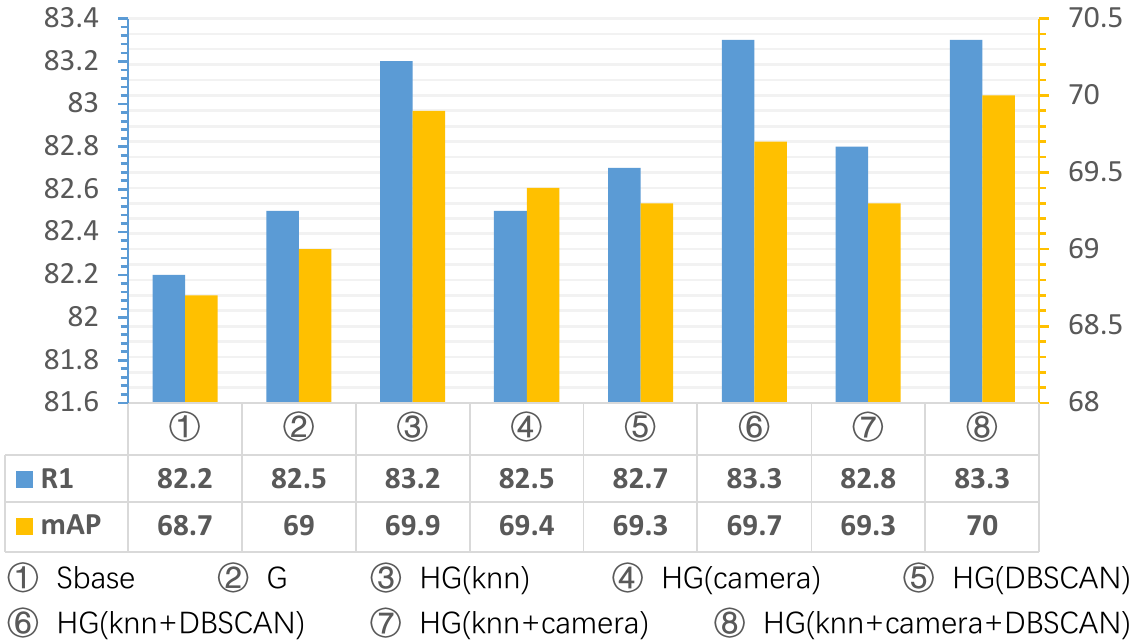}
  \caption{Ablation study on the effectiveness of hypergraph construction on DukeMTMC. Base on \textbf{Sbase}, we investigate different ways to construct hypergraph. KNN is constructing hypergraph with k nearest neighbors, DBSCAN means constructing hypergraph with cluster results of DBSCAN. camera means camera-aware DBSCAN which performs DBSCAN on each camera. Experimental results on Market1501 could be found in the supplementary material.}
  \Description{Ablation study for hypergraph.}
  \label{fig:ablation on hypergraph}
\end{figure}

\vspace{0.2em}
\noindent\textbf{Implementation Details.} We follow same settings as~\cite{luo2019bag} for backbone, which sets the stride of the last residual block as 1 and adds a batch normalization layer. In label generation, we set $\Delta_t$ to 100 frames for spatio-temporal distribution, and we use jaccard distance metric with k-reciprocal nearest neighbors~\cite{zhong2017re} for clustering, $\alpha_1$ and $\alpha_2$ are 0.99 and 0.9 respectively. 
For memory-based supervision, the most hyperparameter setting follows~\cite{wang2020camera}, the instance-level memory is updated by replacing the corresponding features, and camera-aware prototype memory is updated with a moving average rate as 0.2, temperature $\tau$ is set as 0.07, loss weight $\lambda_1, \lambda_2, \lambda_3$ are 1, 1, and 10 separately. %
In the training stage, we adopt random erasing~\cite{zhong2020random}, random flip, cropping as data augmentation strategy, and the images are resized to $256\times 128$ before processing. We set the mini-batch as 32 and use Adam optimizer to optimize the network for 50 epochs, and there are 400 iterations in each epoch. The initial learning rate is 3.5e-4 and divided by 10 every 20 epochs, weight decay is set to 3e-4. In the first 5 epochs, only a weighted intra-camera loss is calculated. In addition, we set $K=10,20,30,40,50$ in ``Global KNN''. And we propose a strong baseline model denoted \textbf{Sbase} based on CAP~\cite{wang2020camera}, which modifies the original intra-camera loss to weighted intra-camera loss and sets the loss weight for inter-camera loss as $1$.

\subsection{Ablation Studies}\label{sec:ablation studies}
\noindent \textbf{Evaluation of Individual Components.} To investigate the effectiveness of individual components in our proposed MGH, e.g. $L_{inst}$ and hypergraph label correction. Here we use Market1501 and DukeMTMC to perform the ablation study, the experimental results are listed in Table~\ref{tab:ablation}. It is apparent that \textbf{Sbase} with weighted intra-camera loss is better than CAP. To be specific, Rank-1 and mAP are improved from 91.5\%/79.7\% to 92.3\%/80.8\% on Market1501 and from 81.5\%/67.3\% to 82.2\%/68.7\% on DukeMTMC. By combining $\L_{inst}$ with \textbf{Sbase}, we could achieve Rank-1/mAP 92.7\%/81.3\% and 83.1\%/69.3\% on Market1501 and DukeMTMC, which indicates the effectiveness of listwise loss based on instance-level memory. In addition, hypergraph label correction could improve Rank-1/mAP by 0.4\%/1.0\% and 1.1\%/1.3\% on these two datasets. Overall, the final model (MGH) could significantly boost the performance by 0.9\%, 0.9\% mAP and 1.4\%, 1.5\% Rank-1 respectively compared with \textbf{Sbase}.

\vspace{0.2em}
\noindent \textbf{The Impact of Hyperparameter $\lambda_3$.}\label{sec:the impact of hyperparameter}
The parameter $\lambda_3$ in Equation~\ref{eq:loss} is crucial to the final performance. In Table~\ref{tab:ablation}, we compare the different values of $\lambda_3$, When $\lambda_3=0$, the method is exactly \textbf{Sbase} that only uses weighted intra-camera loss and inter-camera loss. It is clear that when utilizing instance-memory-based listwise loss, our approach could improve the \textbf{Sbase} at all values. In our experiments, when $\lambda_3$ is set to 10, we achieve the highest results. Thus, we use $\lambda_3=10$ in our final model.

\vspace{0.2em}
\noindent \textbf{Effectiveness of Hypergraph Construction.}\label{sec:effectiveness of hypergraph construction}
As mentioned in Section~\ref{sec:label generation}, how to construct hypergraph is important for modeling data correlation, so we investigate the different hypergraph construction methods. As shown in Figure~\ref{fig:ablation on hypergraph}, we utilize K nearest neighbor (KNN), DBSCAN, and camera-aware DBSCAN (denoted as camera, perform DBSCAN on each camera) to construct hypergraph. Except for the hypergraph, we also present the results of the conventional graph (\textbf{Sbase} w/ G), it also improve the performance compared with \textbf{Sbase}. While the high-order information could not be well modeled in the conventional graph, so the improvement is minor. For KNN, each vertex is treated as the centroid to find the K nearest neighbors, and the neighbors are grouped to generate a hyperedge. For DBSCAN and camera-aware DBSCAN, each cluster is a hyperedge in the hypergraph. The reason why HG(KNN) is superior to HG(KNN+Camera) maybe is the noisy and unstable pseudo labels in “Global Clustering” and “Intra-camera Clustering”. In our experiments, several hyperedge construction strategies are adopted, we find multi-scale KNN is most effective and adopting “Global Clustering” and “Intra-camera Clustering” in hypergraph construction is not stable. From the experimental results, we find combining these three hypergraph construction strategies yields the best performance. Thus, we use HG(KNN+camera+DBSCAN) to build a hypergraph in our proposed method. 

According to the experimental results, in our opinion, constructing hypergraph should follow three criterions in unsupervised person ReID: 1) Complementary. different strategies should reinforce each other. 2) Accurate and stable. The information for hypergraph construction should be accurate for modelling the data correlation. 3) Multi-scale. Our experimental results indicate that multi-scale local information is effective.

\subsection{Comparison with State-of-the-art Methods}
In this section, we compare our proposed MGH with the state-of-the-art unsupervised and domain adaptive person ReID approaches on Market1501, DukeMTMC, and MSMT17 datasets.

\vspace{0.2em}
\noindent \textbf{Comparison on Market1501 and DukeMTMC.} We compare our proposed method with the state-of-the-are unsupervised learning approaches, including 15 approaches without meta information~\cite{lin2019bottom,ge2020self,zeng2020hierarchical,wang2020unsupervised,lin2020unsupervised,wang2020cycas,zhang2021refining,fan2018unsupervised,deng2018image,fu2019self,song2020unsupervised,ge2020mutual,zhai2020multiple,zhai2020ad,zheng2020exploiting}, and 5 approaches with meta information~\cite{li2020joint,zhong2019invariance,zhong2020learning,wang2020camera,lv2018unsupervised,wu2019unsupervised}. Table~\ref{tab:comparison with Market and DukeMTMC} shows the results on Market1501 and DukeMTMC, we observe that our proposed method outperforms all the previous approaches. Compared with the best unsupervised ReID approach SpCL~\cite{ge2020self}, MGH outperforms by a large margin. Even compared with the UDA methods which exploits an external labeled data, we still have a better result. In comparison with the best UDA methods UNRN~\cite{zheng2020exploiting}, although we only use data in the target domain, we still obtain an improvement of 1.3\% Rank-1 and 3.6\% on Market1501, 1.7\% Rank-1 and 1.1\% mAP on DukeMTMC. When only target dataset is accessible, our method surpasses the best method CAP~\cite{wang2020camera} by 1.8\% Rank-1 and 2.5\% mAP on Market1501, 2.6\% Rank-1 and 2.9\% mAP on DukeMTMC. 
MGH+ could improve the performance further by using joint similarity to compute the query-gallery similarity in testing stage.

\begin{table}
  \centering
  \begin{tabular}{lccccc} 
  \hline
  \multirow{2}{*}{Method}                                                 & \multicolumn{5}{c}{MSMT17}           \\ 
  \cline{2-6}
                                                                          & source  & R1   & R5   & R10  & mAP   \\ 
  \hline
  \multicolumn{6}{l}{\textcolor[rgb]{0.502,0.502,0.502}{Unsupervised
    person ReID}}                                                                 \\ 
  \hline
  SpCL~\cite{ge2020self}                 & None    & 42.3 & 55.6 & 61.2 & 19.1  \\
  MMCL~\cite{wang2020unsupervised}       & None    & 35.4 & 44.8 & 49.8 & 11.2  \\
  \hline
  \multicolumn{6}{l}{\textcolor[rgb]{0.502,0.502,0.502}{UDA person ReID}}                                                                          \\ 
  \hline
  MMCL~\cite{wang2020unsupervised}       & Market  & 40.8 & 51.8 & 56.7 & 15.1  \\
  MMT~\cite{ge2020mutual}                & Market  & 50.1 & 63.5 & 69.3 & 24.0  \\
  SpCL~\cite{ge2020self}                 & Market  & 51.6 & 64.3 & 69.7 & 25.4  \\
  DIM+GLO~\cite{liu2020domain}                        & Market  & 49.7 & -    & 66.1 & 20.7  \\
  UNRN~\cite{zheng2020exploiting}        & Market  & 52.4 & 64.7 & 69.7 & 25.3  \\
  MMCL~\cite{wang2020unsupervised}         & Duke    & 43.6 & 54.3 & 58.9 & 16.2  \\
  MMT~\cite{ge2020mutual}                & Duke    & 52.9 & 66.3 & 71.3 & 25.1  \\
  SpCL~\cite{ge2020self}                 & Duke    & 53.1 & 65.8 & 70.5 & 26.5  \\
  DIM+GLO~\cite{liu2020domain}           & Duke    & \textcolor{blue}{56.5} & -    & 70.0 & 24.4  \\ 
  UNRN~\cite{zheng2020exploiting}        & Duke    & 54.9 & 67.3 & 70.6 & 26.2  \\ 
  \hline
  \multicolumn{6}{l}{\textcolor[rgb]{0.502,0.502,0.502}{Unsupervised person ReID with meta information}}                                           \\ 
  \hline
  UGA~\cite{wu2019unsupervised}          & Tacklet & 49.5 & -    & -    & 21.7  \\
  CAP~\cite{wang2020camera}              & None    & \textcolor{green}{67.4} & \textcolor{green}{78.0} & \textcolor{green}{81.4} & \textcolor{green}{36.9}  \\
  ECN++~\cite{zhong2020learning}         & Market  & 40.4 & 53.1 & 58.7 & 15.2  \\
  JVTC+~\cite{li2020joint}               & Market  & 48.6 & 65.3 & 68.2 & 25.1  \\

  ECN++~\cite{zhong2020learning}         & Duke    & 42.5 & 55.9 & 61.5 & 16.0  \\
  JVTC+~\cite{li2020joint}               & Duke    & 52.9 & \textcolor{blue}{70.5} & \textcolor{blue}{75.9} & \textcolor{blue}{27.5}  \\  
  \hline
  \multicolumn{1}{l}{MGH(Ours)}        & None    & \textcolor{red}{70.2} & \textcolor{red}{81.2} & \textcolor{red}{84.5} & \textcolor{red}{40.6}  \\
  \multicolumn{1}{l}{MGH+(Ours)}       & None    & 75.2 & 83.9 & 86.7 & 43.6  \\
  \hline
  \end{tabular}
  \caption{Comparison with state-of-the-are methods on MSMT17. The top-3 best results are highlighted with \textcolor{red}{red}, \textcolor{green}{green}, and \textcolor{blue}{blue} respectively. MGH+ uses joint similarity to compute the query-gallery similarity.}
  \label{tab:comparison with MSMT17}
  \vspace{-1.5em}
\end{table}

\vspace{0.2em}
\noindent \textbf{Comparison on MSMT17.} On MSMT17, we compare our MGH with several methods and report the results on Table~\ref{tab:comparison with MSMT17}. The results obtained by MGH are 70.2\%/40.6\% on Rank-1/mAP accuracy, which exceeds the second best method CAP by 2.8\%/3.7\% on Rank-1/mAP. Compared with the supervised ReID method PCB, our method has a better result. This demonstrates the effectiveness of our proposed MGH on the large scale dataset.

\begin{table}
  \centering
  \begin{tabular}{lrrrrrr} 
  \hline
  \multicolumn{1}{c}{\multirow{2}{*}{Method}} & \multicolumn{2}{c}{Market1501}                   & \multicolumn{2}{c}{DukeMTMC}                     & \multicolumn{2}{c}{MSMT17}                        \\ 
  \cline{2-7}
  \multicolumn{1}{c}{}                        & \multicolumn{1}{c}{R1} & \multicolumn{1}{c}{mAP} & \multicolumn{1}{c}{R1} & \multicolumn{1}{c}{mAP} & \multicolumn{1}{c}{R1} & \multicolumn{1}{c}{mAP}  \\ 
  \hline
  PCB~\cite{sun2018beyond}                                         & 93.8                   & 81.6                    & 83.3                   & 69.2                    & 68.2                   & 40.4                     \\
  BoT~\cite{luo2019bag}                                         & 94.5                   & 85.9                    & 86.4                   & 76.4                    & 74.1                   & 50.2                     \\
  ABD-Net~\cite{chen2019abd}                                      & 95.6                   & 88.3                    & 89                     & 78.6                    & 82.3                   & 60.8                     \\\hline
  MGH                                         & 93.2                   & 81.7                    & 83.7                   & 70.2                    & 70.2                   & 40.6                     \\
  MGH+                                        & 95.5                   & 84                      & 91                     & 77.2                    & 75.2                   & 43.6                     \\
  \hline
  \end{tabular}
  \caption{Comprison with supervised person ReID methods on three benchmarks.}
  \label{tab:comparison with supervised}
  \vspace{-1.5em}
\end{table}

\vspace{0.2em}
\noindent \textbf{Comparison with Supervised Approaches.} We further present the 
comparison with the supervised counterpart. As shown in Table~\ref{tab:comparison with supervised}, our proposed unsupervised method MGH even outperforms PCB, which is trained with labeled information. And the gap with the state-of-the-art supervised method ABD-Net has been mitigated on three benchmarks. By further using joint similarity in testing stage, MGH+ can greatly narrow down the margin between ABD-Net.

\section{Conclusion}
This paper tackles the unsupervised person Re-ID by jointly considering the rich auxiliary meta information attached with the captured images in the distributed camera network. By constructing hypergraphs, we model the high-order data correlation and encode the multi-modal information, then a label correction approach is utilized to refine the noisy pseudo labels. Furthermore, to directly optimize AP, we propose a memory-based listwise loss, which stores the instance features in the memory bank, and used as the gallery in the training stage. Finally, to bridge the gap between experimental settings and real-world ReID tasks, we validate the effectiveness of our method on three widely used benchmarks and show our proposed MGH achieves a significant improvement compared with the previous approaches.

\begin{acks}
  This work is supported in part by National Key Research and Development Program of China under Grant 2020AAA0107400, Zhejiang Provincial Natural Science Foundation of China under Grant LR19F020004, key scientific technological innovation research project by Ministry of Education, and National Natural Science Foundation of China under Grant U20A20222, and Collaborative Innovation Center of Artificial Intelligence by MOE and Zhejiang Provincial Government (ZJU).
\end{acks}

\bibliographystyle{ACM-Reference-Format}
\bibliography{wu}
\end{document}